\documentclass[conference]{IEEEtran}
\IEEEoverridecommandlockouts
\usepackage{cite}
\usepackage{amsmath,amssymb,amsfonts}
\usepackage{algorithmic}
\usepackage{graphicx}
\usepackage{textcomp}
\usepackage{xcolor}
\usepackage{booktabs}
\usepackage{multirow}
\usepackage{subcaption}
\usepackage{hyperref}
\usepackage{url}
\def\BibTeX{{\rm B\kern-.05em{\sc i\kern-.025em b}\kern-.08em
    T\kern-.1667em\lower.7ex\hbox{E}\kern-.125emX}}
\begin{document}

\title{COMPASS: COmpact Multi-channel Prior-map And Scene Signature for Floor-Plan-Based Visual Localization}
\author{Muhammad Shaheer, Miguel Fernandez-Cortizas, Asier Bikandi-Noya, \\ Holger Voos and Jose Luis Sanchez-Lopez  
\thanks{Authors are with the Automation and Robotics Research Group, Interdisciplinary Centre for Security, Reliability and Trust (SnT), University of Luxembourg (UL). Holger Voos is also associated with the Faculty of Science, Technology and Medicine, University of Luxembourg, Luxembourg.
\tt{\small{\{muhammad.shaheer, miguel.fernandez, asier.bikandi, holger.voos, joseluis.sanchezlopez\}}@uni.lu}}%
\thanks{*This work was partially funded by the Fonds National de la Recherche of Luxembourg (FNR) under the project 19685965/BARCODE.
For Open Access, the author has applied a CC BY 4.0 public copyright license to any Author Accepted Manuscript version arising from this submission.}
}
\maketitle
\begin{abstract}
 Architectural floor plans are widely available priors which contain not only geometry but also the semantic information of the environment, yet existing localization methods largely ignore this semantic information. To address this, we present COMPASS, an algorithm that exploits both geometric and semantic priors from floor plans to estimate the pose of a robot equipped with dual fisheye cameras. Inspired by scan context descriptor from LiDAR-based place recognition, we design a multi-channel radial descriptor that encodes the geometric layout surrounding a position. From the floor plan, rays are cast in 360 azimuth bins and the results are encoded into five channels: normalized range, structural hit type (wall, window, or opening), range gradient, inverse range, and local range variance. From the image side, the same descriptor structure is populated by detecting structural elements in the fisheye imagery. As a first step toward full cross-modal matching, we present a window detection algorithm for fisheye images that uses a line segment detector to identify window frames via vertical edge clustering and brightness verification. Detected windows are projected to azimuthal bearings through the fisheye camera model, producing the hit-type channel of the visual descriptor. As a proof of concept, we generate both descriptors at a single known pose from the Hilti-Trimble SLAM Challenge 2026 dataset and demonstrate that the wall-window pattern extracted from the first frame of each camera closely matches the floor plan descriptor, validating the feasibility of cross-modal structural matching.
\end{abstract}
\section{Introduction}
Indoor localization remains a core challenge for mobile robots in service, logistics, and 
construction applications. While LiDAR-based SLAM systems have matured considerably~\cite{cadena2016past}, they rely on expensive range sensors and must build maps
from scratch in every new environment. At the same time, 2D architectural floor plans are 
routinely produced during building design and encode the geometric layout of walls, doors, and windows with sufficient accuracy for navigation. 
\begin{figure*}[!t]
    \centering
\includegraphics[width=\textwidth,height=0.42\textheight,keepaspectratio]{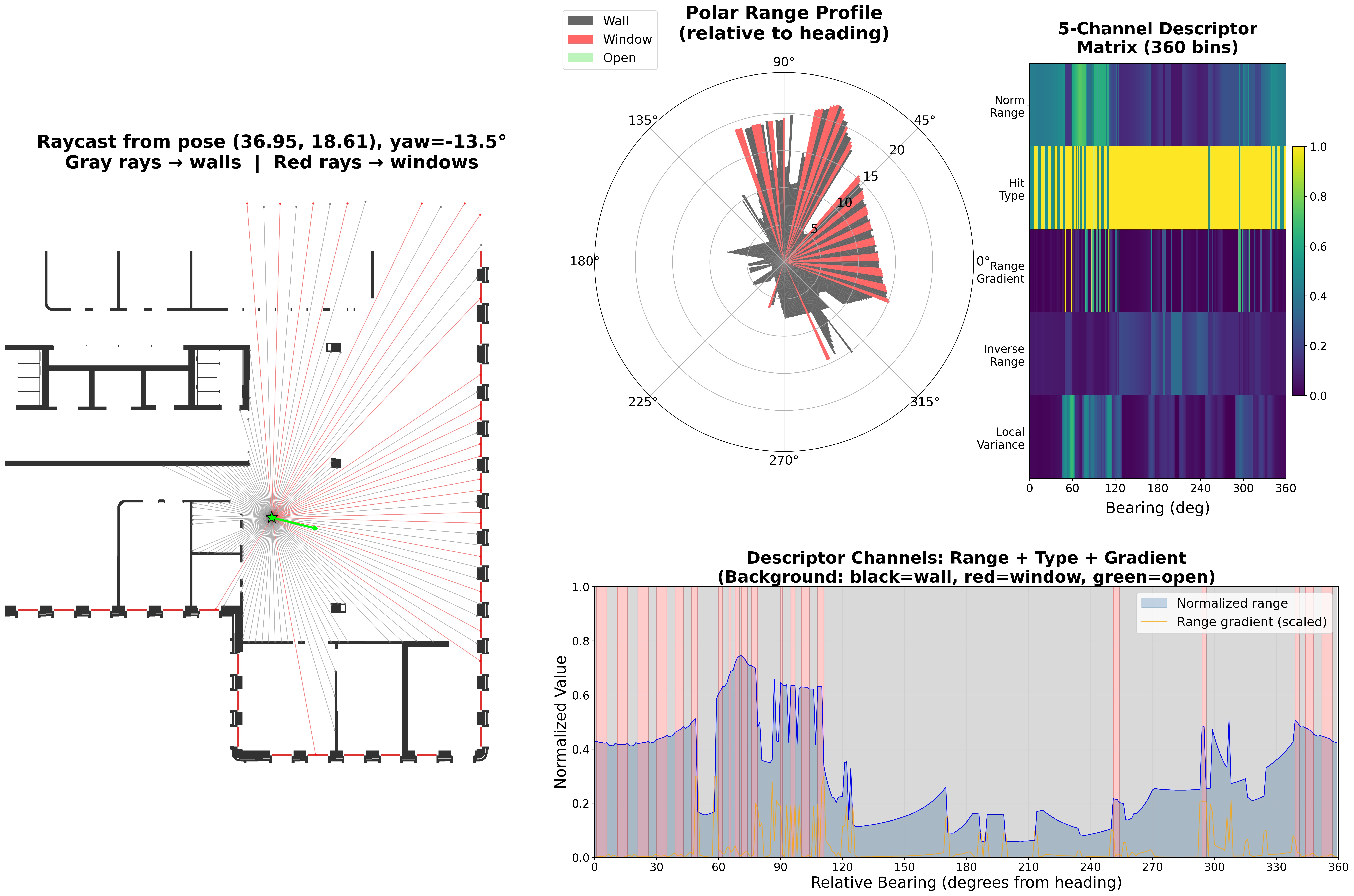}
    \caption{Floor plan descriptor. \textbf{Top left}: ray cast overlaid on the floor plan (gray rays $\to$ walls, red $\to$ windows). \textbf{Top center}: polar range profile. \textbf{Top right}: the $5 \times 360$ descriptor matrix. \textbf{Bottom}: linearized range (blue), gradient (orange), with background shading by hit type.}
    \label{fig:descriptor}
\end{figure*}
Existing floor-plan-based localization methods, however, underutilize the semantic information that floor plans provide. Particle-filter approaches~\cite{boniardi2019robot, winterhalter2015accurate} treat all structural boundaries identically, ignoring the distinction between walls and windows. LiDAR-based graph methods~\cite{shaheer2023graph, shaheer2025tightly} exploit higher-level semantics but require 3D range sensors. Scan Context~\cite{kim2018scan} and its variants~\cite{kim2021scan, wang2020intensity} are effective for LiDAR place recognition, but encode only geometric properties (primarily height). They do not encode semantic element types, which limits discriminative power in indoor environments where rooms with similar geometry differ in their wall-window layout.

We introduce \textbf{COMPASS} (\textbf{CO}mpact \textbf{M}ulti-channel \textbf{P}rior-map \textbf{A}nd \textbf{S}cene \textbf{S}ignature), a descriptor for global localization that explicitly encodes both geometry and structural semantics in a unified representation. Our key insight is that the alternating pattern of walls, windows, and openings surrounding any indoor position constitutes a highly discriminative signature, and this pattern is encoded in the floor plan and visually observable in camera images. We propose a \emph{multi-channel radial descriptor} of shape $C \times N_s$ (with $C{=}5$ channels and $N_s{=}360$ azimuth bins at $1^{\circ}$ resolution) that captures range, \emph{structural element type} (wall, window, or opening), range gradient, inverse range, and local range variance. By jointly encoding what an element is and where it is, COMPASS produces a more discriminative descriptor than purely geometric approaches.


Our contributions are: (1)~a five-channel radial descriptor that jointly encodes geometry and structural semantics from a 2D floor plan, going beyond purely geometric descriptors; (2)~a window detection and bearing estimation algorithm for fisheye images that produces the hit-type channel of the same descriptor; and (3)~cross-modal comparison demonstrating pattern agreement between floor plan and image descriptors.
\section{COMPASS based Localization}
COMPASS consists of three components: (A)~floor plan descriptor generation via ray casting, (B)~visual descriptor generation from dual fisheye images, and (C)~pose recovery, comprising descriptor matching for position and yaw via circular cross-correlation, and roll/pitch estimation via vanishing point analysis.
\subsection{Floor Plan Descriptor Generation}\label{floor_plan_descriptor}
The floor plan is represented as two binary raster masks (see Fig. \ref{fig:descriptor}) at metric resolution $\rho$ (m/pixel): a \emph{wall mask} $\mathcal{M}_w$ (dark pixels indicate walls) and a \emph{window mask} $\mathcal{M}_g$ (red-colored pixels mark glazing).
%
%
\subsubsection{Ray Casting}
From a candidate position $\mathbf{p} = (t_x, t_y)$ and heading $\psi$, we cast $N_s = 360$ rays at uniformly spaced relative bearings $\theta_j = 2\pi j / N_s$. Each ray is marched at step size $\Delta_s$ until hitting a wall pixel, a window pixel, or reaching $r_{\max}$. This yields a range $r_j$ and hit-type label $h_j \in \{\text{wall}, \text{window}, \text{open}\}$ per bin.
\subsubsection{Five-Channel Descriptor}
The outputs are encoded into $\mathbf{D}^{\text{map}} \in \mathbb{R}^{5 \times N_s}$:

\textbf{Ch.\,0---Normalized range:} $\mathbf{D}(0, j) = r_j / r_{\max}$. 
It distinguishes positions that are close to a wall from those in the middle of a large room. Two positions at different distances from the same wall will differ in this channel even if they share the same wall--window pattern.

\textbf{Ch.\,1---Hit type:} $\mathbf{D}(1, j) = 1.0$ (wall), $0.5$ (window), or $0.0$ (open). This channel encodes the alternating wall--window pattern characteristic of building facades.

\textbf{Ch.\,2---Range gradient:} $\mathbf{D}(2, j) = \min(|\nabla_\theta r_j| / r_{\text{clip}},\, 1)$, highlighting corners and depth discontinuities (clipped at $r_{\text{clip}} = 5$\,m). The range is usually smooth along a flat wall but jumps sharply at corners, doorways, and column edges where the range to the nearest surface changes abruptly. 

\textbf{Ch.\,3---Inverse range:} $\mathbf{D}(3, j) = 1/(1 + r_j)$, correlating with the angular extent of elements in fisheye images. A wall at 2\,m distance occupies a much larger angular extent in a fisheye image than the same wall at 20\,m. Inverse range captures this relationship. Nearby surfaces produce large values, distant ones produce small values.

\textbf{Ch.\,4---Local range variance:} $\mathbf{D}(4, j) = \text{std}(\{r_k : k \in \mathcal{N}_w(j)\}) / \sigma_{\text{clip}}$, capturing structural complexity within a window of half-width $w{=}5$ bins ($\sigma_{\text{clip}} = 10$\,m). Some positions are surrounded by smooth, featureless walls (low variance), while others are near structurally complex areas such as corridor entrances, columns, or room boundaries, where the range changes rapidly across nearby bearings (high variance). This channel captures that structural complexity at a local scale, providing a complementary cue to the point-wise gradient of Ch.\,2.

Fig.~\ref{fig:descriptor} visualizes an example descriptor.
\subsection{Visual Descriptor from Dual Fisheye Images}\label{fvisual_descriptor}
The Hilti-Trimble 2026 dataset~\cite{hilti2026} provides dual fisheye images (front and back) from an Insta360 camera, jointly covering ${\sim}360^{\circ}$. The goal is to generate a visual descriptor $\mathbf{D}^{\text{vis}} \in \mathbb{R}^{5 \times N_s}$ using the same channel encoding as the floorplan descriptor. Because both descriptors encode the same physical structure in the same format, they can be directly compared via cross-correlation. To populate $\mathbf{D}^{\text{vis}}$, Ch.\,1 (hit type) is filled by detecting windows in the fisheye images and projecting them to azimuth bins. Bins with detected windows are set to $0.5$, all others default to $1.0$ (wall). This is the channel we validate in this work. Ch.\,0 (range) can in principle be estimated from the floor--wall boundary position in the fisheye image and the known camera height; this channel and its derivatives (Ch.\,2--4) are the subject of ongoing work.
\subsubsection{Window Detection via ELSED and Vertical Edge Clustering}

We detect windows in each fisheye image through a segment-first approach 

\textbf{Step 1: Line segment detection.} ELSED~\cite{suarez2022elsed} is applied to detect line segments, producing ${\sim}$200--400 validated segments per image. ELSED's edge-drawing approach is well-suited to the low-texture concrete walls typical of construction environments.

\textbf{Step 2: Window band estimation.} The vertical region of the image where windows concentrate is auto-detected by combining a brightness density profile (bright pixels indicate glass/opening) with vertical segment density, yielding a band $[y_{\text{top}}, y_{\text{bot}}]$.

\textbf{Step 3: Vertical segment extraction and clustering.} Segments within the window band that are more than $30^{\circ}$ from horizontal and longer than a minimum threshold are selected. Nearby vertical segments are clustered (within a horizontal gap of 8\,px and requiring vertical overlap) to form single edge hypotheses.

\textbf{Steps 4--5: Pairing and filtering.} Pairs of edge clusters with appropriate separation and vertical overlap form window proposals, verified by interior brightness, wall contrast, and texture non-uniformity. Non-maximum suppression removes overlapping detections, and additional filters reject safety equipment (red-dominant regions) and fisheye-periphery artifacts.
\begin{figure}[!h]
    \centering
    \begin{subfigure}[t]{0.20\textwidth}
        \centering
        \includegraphics[width=\textwidth]{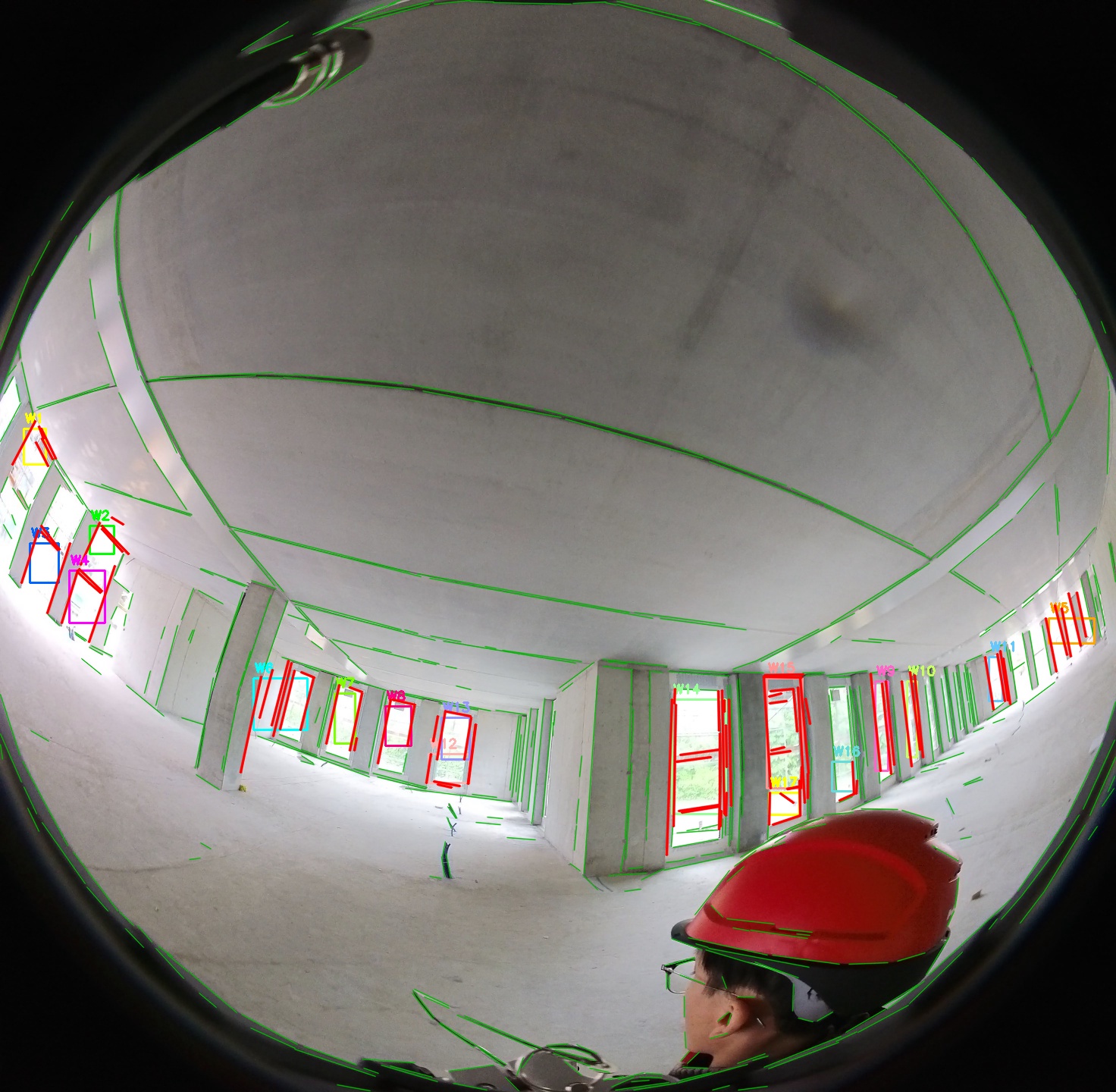}
        \caption{Front camera}
        \label{fig:fisheye_front}
    \end{subfigure}
    \hfill
    \begin{subfigure}[t]{0.20\textwidth}
        \centering
        \includegraphics[width=\textwidth]{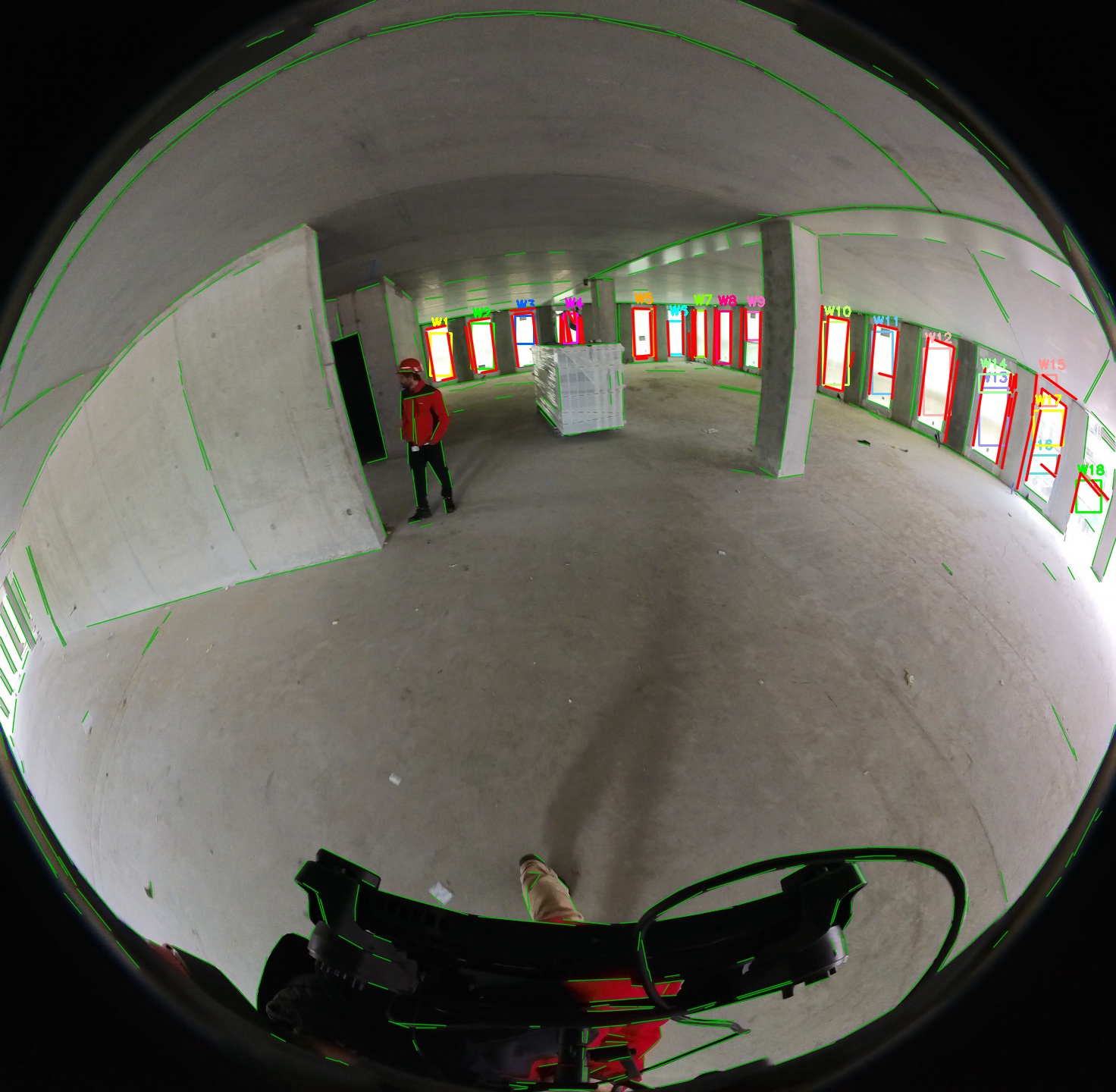}
        \caption{Back camera}
        \label{fig:fisheye_back}
    \end{subfigure}
    \caption{Window detection on dual fisheye images from a construction site. Green lines: ELSED line segments. Red lines: segments associated with detected windows.}
    \label{fig:fisheye}
\end{figure}
\begin{figure*}[!t]
    \centering
    \begin{subfigure}[t]{0.35\textwidth}
        \centering
        \includegraphics[width=\textwidth,height=0.22\textheight,keepaspectratio]{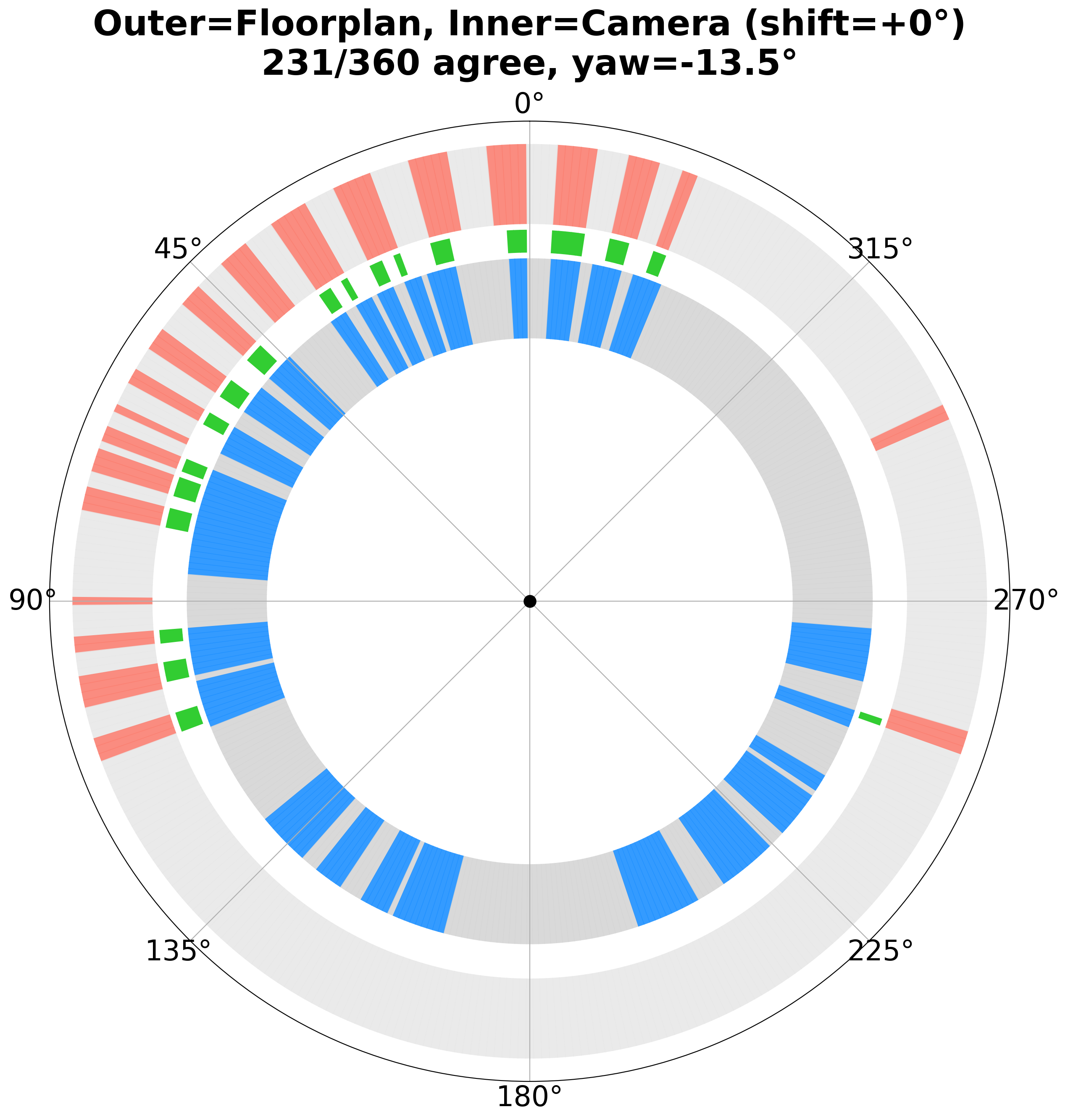}
        \caption{Polar overlay of hit type channel. Outer: floor plan, Inner: camera. Green: both agree on the window.}
        \label{fig:hittype_polar}
    \end{subfigure}
    \hfill
    \begin{subfigure}[t]{0.42\textwidth}
        \centering
        \includegraphics[width=\textwidth]{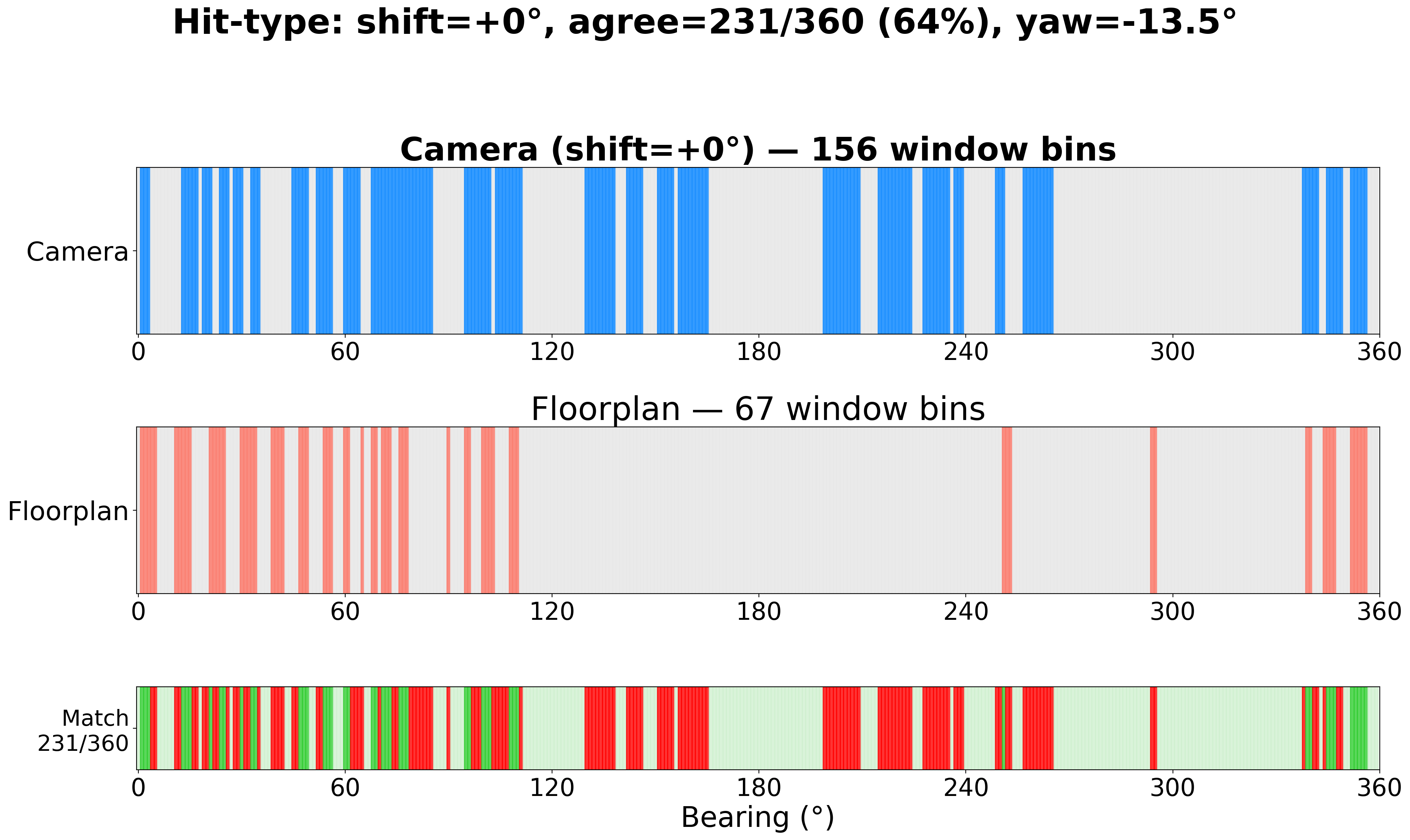}
        \caption{Linear comparison. Top: camera (blue\,=\,window). Middle: floor plan (red\,=\,window). Bottom: agreement (green) / disagreement (red).}
        \label{fig:hittype_linear}
    \end{subfigure}
    \caption{Cross-modal hit-type matching. The camera hit-type (156 window bins) is compared against the floor plan (67 window bins) at the ground-truth pose. 
    }
    \label{fig:hittype}
\end{figure*}
\subsubsection{Window Bearing Estimation}\label{sec:bearing}
Each detected window bounding box $(b_x, b_y, b_w, b_h)$ is projected to an azimuthal bearing using the Kannala--Brandt fisheye model~\cite{kannala2006generic}. The center pixel is unprojected to a 3D bearing vector:
\begin{equation}
    \mathbf{b} = \begin{bmatrix} \sin\theta\cos\phi \ \sin\theta\sin\phi \  \cos\theta \end{bmatrix}^T
    \label{eq:bearing}
\end{equation}
where $\theta = \min(r/f, \theta_{\max})$ is the incidence angle (with $r$ the pixel distance from the principal point and $f$ the focal length), and $\phi = \text{atan2}(dy, dx)$ is the pixel direction. For back-camera detections, the bearing is rotated by $180^{\circ}$ via $\mathbf{b}_{\text{back}} = \mathbf{R}_\pi \mathbf{b}$. The azimuth is extracted as $\alpha = \text{atan2}(b_x, b_z)$.

The angular span of each window is computed from its left and right edges, and the corresponding bins in a 360-element hit-type array are set to $0.5$ (window); all remaining bins default to $1.0$ (wall).
\subsection{Descriptor Matching and Pose Recovery}\label{sec:matching}
A key property of the COMPASS descriptor, inspired by Scan Context \cite{scancontext} is \emph{rotation equivariance}: a heading change $\Delta\psi$ corresponds to a cyclic column shift of $\Delta n = \lfloor \Delta\psi \cdot N_s / 2\pi \rceil$. Given a query $\mathbf{D}^{\text{vis}}$ and a database of precomputed floor plan descriptors $\mathcal{D} = \{\mathbf{D}^{\text{map}}_k\}$, the best-matching position and heading are found by maximizing the cosine similarity over all candidates and column shifts, computed efficiently via FFT-based circular cross-correlation at cost $O(C \cdot N_s \log N_s)$ per candidate. The transition signature provides optional coarse pre-filtering.
\subsection{Roll and Pitch Estimation}\label{sec:rollpitch}
Roll and pitch are decoupled from the planar pose. Vertical lines detected by ELSED provide cues for gravity estimation via vanishing point analysis under the fisheye model~\cite{lochman2021minimal, antunes2017unsupervised}, from which roll and pitch are extracted via RANSAC.
\section{Preliminary Results}\label{sec:results}
We present a single-pose proof of concept using the Hilti-Trimble SLAM Challenge 2026 dataset~\cite{hilti2026}. The test environment is a ${\sim}40 \times 30$\,m, still under construction with floor plan at $0.01$\,m/pixel (Fig.~\ref{fig:descriptor}). Using the ground-truth 6-DoF starting pose, we generate the floor-plan descriptor and extract the image-side descriptor from frames of each Insta360 camera ($1472 \times 1440$, ${\sim}190^{\circ}$ FoV) captured at the same spot.

\noindent \textbf{Floor plan descriptor.} Using $N_s = 360$, $r_{\max} = 30$\,m, and $\Delta_s = 0.02$\,m, the descriptor at the starting pose captures ranges from 1.76 to 22.36\,m with 43 wall--window segments (67 window bins out of 360). The range gradient channel is sparse (mean 0.09), providing discriminative features at corners.

\begin{figure}[h]
    \centering
    \includegraphics[width=0.75\columnwidth]{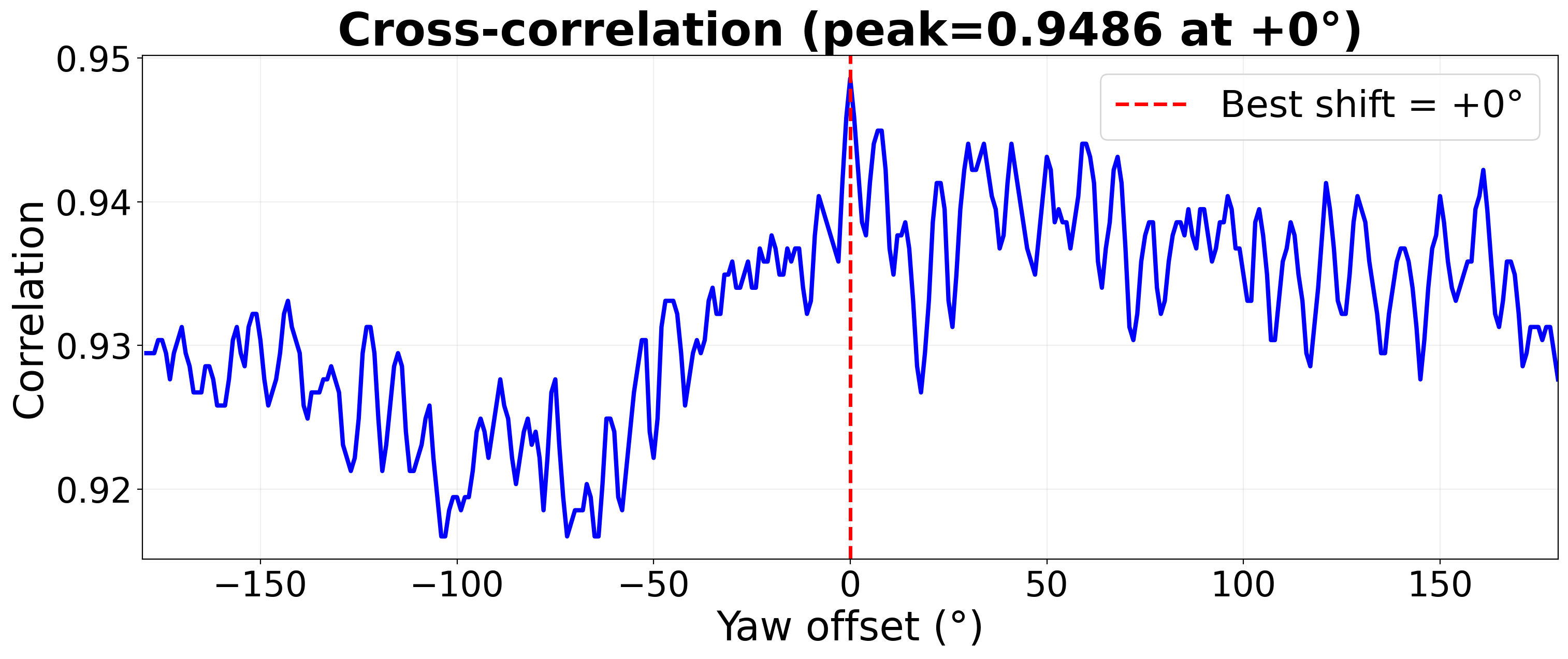}
\caption{Cross-correlation peaks at $0^{\circ}$ (score 0.9486).}
        \label{fig:xcorr}
\end{figure}

\noindent \textbf{Window detection and hit-type matching.} From the front and back fisheye frames, ELSED \cite{suarez2022elsed} detects ${\sim}$300 segments per image. The window extraction algorithm identifies 18 and 16 windows, respectively, yielding 156 window bins in the camera hit-type channel. We perform FFT-based circular cross-correlation between the camera and floor plan hit-type channels (Fig.~\ref{fig:xcorr}). The correlation peaks at 0.9486 with a shift of $0^{\circ}$, correctly recovering the relative heading.
At this alignment, 231 out of 360 bins (64\%) agree between the two modalities (Fig.~\ref{fig:hittype}). The main disagreements are between $110^{\circ}$--$150^{\circ}$, where the camera detects windows on a distant wall the floor plan encodes as a single segment, and $180^{\circ}$--$280^{\circ}$, where the camera observes windows behind a wall that had not yet been constructed at capture time which highlights a key challenge of construction-site localization where the as-built state deviates from the plan. Despite these discrepancies, cross-correlation produces a clear peak at the correct heading.
\begin{figure}[!b]
    \centering
    \begin{subfigure}[t]{0.4\columnwidth}
        \centering
        \includegraphics[width=\textwidth]{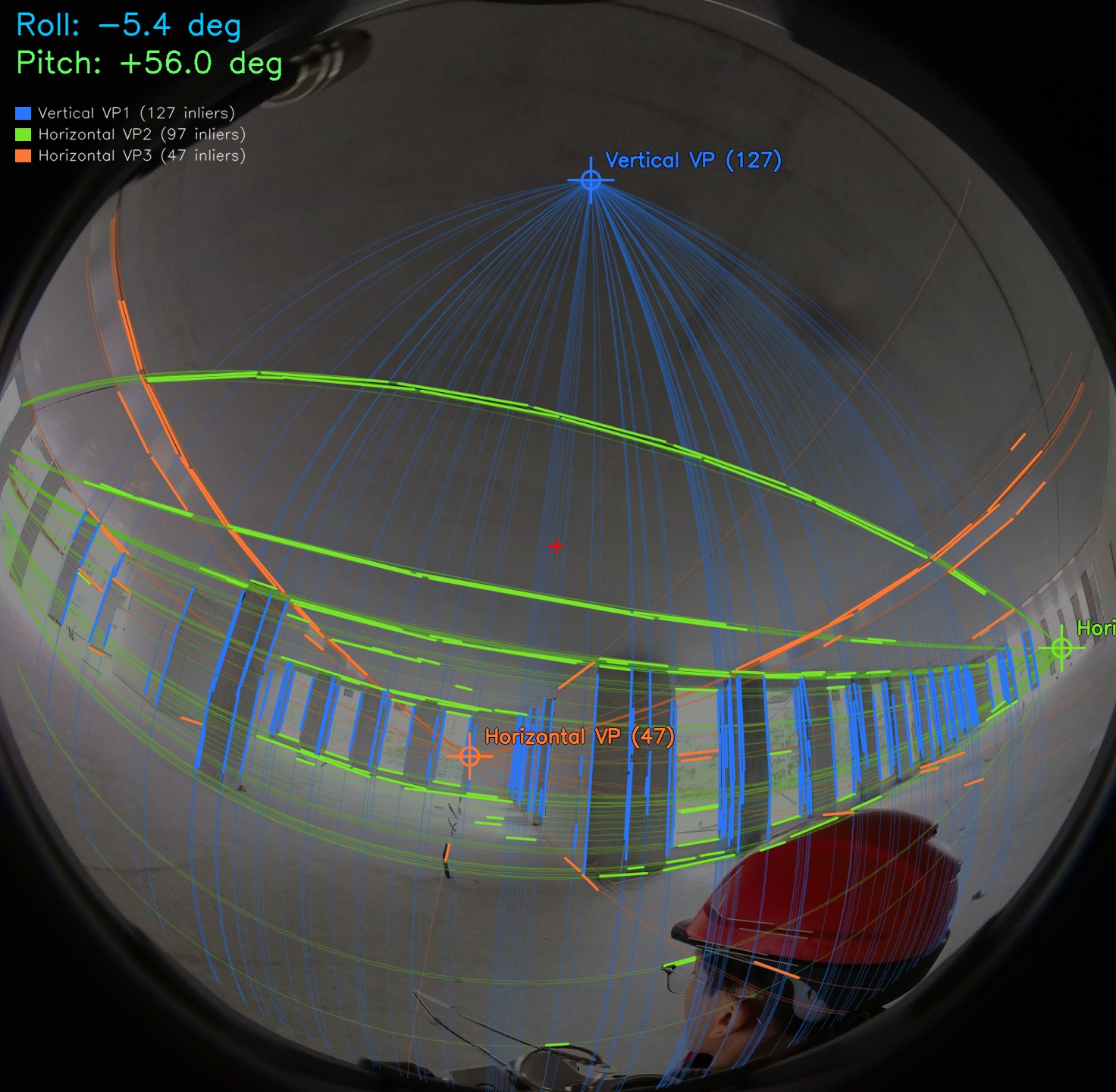}
        \caption{Front camera: vertical VP with 127 inliers (blue), two horizontal VPs (green: 97, orange: 47 inliers).}
    \end{subfigure}
    \hfill
    \begin{subfigure}[t]{0.4\columnwidth}
        \centering
        \includegraphics[width=\textwidth]{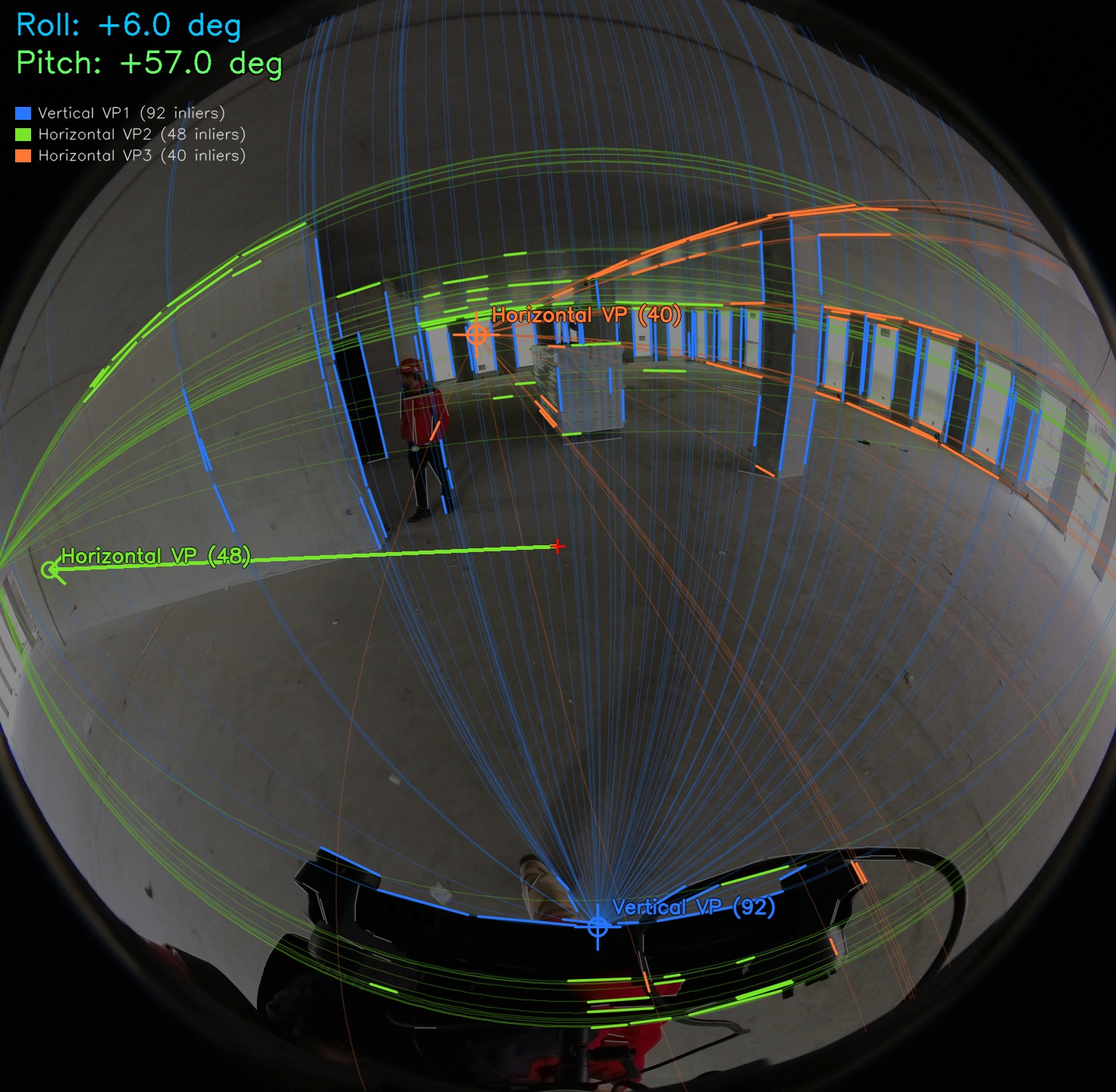}
        \caption{Back camera: vertical VP with 92 inliers (blue), two horizontal VPs (green: 48, orange: 40 inliers).}
    \end{subfigure}
    \caption{Vanishing point estimation for roll and pitch recovery. Blue segments converge to the vertical VP (gravity direction), green and orange to horizontal VPs corresponding to building axes.}
    \label{fig:vp}
\end{figure}
\textbf{Roll and pitch estimation.} The Vanishing point (VP) based attitude estimation algorithm is applied to the same pair of fisheye frames. Fig.~\ref{fig:vp} visualizes the detected VPs and their inlier segments for both cameras, with great-circle arcs showing how line segments converge toward each VP on the unit sphere.

\textbf{Limitations and ongoing work.}
While the hit-type channel alone recovers the correct heading at this test pose, it is insufficient for general localization. First, this pose lies near a window-rich facade, but many indoor positions (interior corridors, rooms away from exterior walls) have few or no visible windows, making the hit-type channel nearly uniform and non-discriminative. Second, hit-type captures \emph{what} is at each bearing but not \emph{how far} it is: multiple candidate positions along the same facade would share a similar wall--window pattern, and range-based channels are essential to disambiguate position. Third, the range gradient and local variance channels encode geometric features (corners, alcoves, corridor entrances) that are present everywhere, even in window-free interiors. In short, a hit-type channel is not enough for robust localization in indoor environments. The hit-type is most discriminative near facades, while range-derived channels provide discrimination in geometrically complex interiors.
\section{Conclusion}\label{sec:conclusion}
We presented COMPASS, a framework for floor-plan-based localization that jointly encodes geometry and structural semantics in a compact, rotation-equivariant $5 \times 360$ radial descriptor. As a proof of concept on the Hilti-Trimble 2026 dataset, we showed that the hit-type channel extracted from a single pair of fisheye frames closely matches the floor plan descriptor, with cross-correlation correctly recovering the heading at $0^{\circ}$ shift. We also demonstrated roll and pitch estimation via stereo vanishing point analysis on the unit sphere, achieving ${\sim}2^{\circ}$ agreement with ground truth on the dominant tilt axis. Together, these results validate that structural semantic patterns and geometric line features in fisheye images can serve as viable signals for floor-plan-based localization. Ongoing work targets completing all five descriptor channels from imagery, implementing full descriptor matching over dense candidate grids, and evaluating pose recovery across entire trajectories and multiple environments.

\end{document}